\documentclass[11pt,a4paper]{article}
\usepackage{authblk}
\usepackage[hyperref]{naaclhlt2019}
\usepackage{times}
\usepackage{latexsym}
\usepackage{booktabs}
\usepackage{multirow}
\usepackage{tikz}
\usepackage{tikz-dependency}
\usepackage{todonotes}
\usepackage{url}
\usepackage{subcaption}

\aclfinalcopy 
\def\aclpaperid{20} 

\title{From News to Medical: Cross-domain Discourse Segmentation}
\author[1]{{\bf Elisa Ferracane}}
\author[2]{{\bf Titan Page}}
\author[1]{{\bf Junyi Jessy Li}}
\author[1]{{\bf Katrin Erk}}
\affil[1]{Department of Linguistics, The University of Texas at Austin}
\affil[2]{Department of Linguistics, University of Colorado Boulder}
\affil[ ]{\tt elisa@ferracane.com, titan@colorado.edu}
\affil[ ]{\tt jessy@austin.utexas.edu, katrin.erk@mail.utexas.edu}

\begin{document}
\maketitle
\begin{abstract}
  The first step in discourse analysis involves dividing a text into segments. We annotate the first high-quality small-scale medical corpus in English with discourse segments and analyze how well news-trained segmenters perform on this domain. While we expectedly find a drop in performance, the nature of the segmentation errors suggests some problems can be addressed earlier in the pipeline, while others would require expanding the corpus to a trainable size to learn the nuances of the medical domain.\footnote{Code and data available at \url{http://github.com/elisaF/news-med-segmentation}.}
\end{abstract}

\section{Introduction}

Dividing a text into units is the first step in analyzing a discourse. In the framework of Rhetorical Structure Theory (RST) \cite{Mann:1988}, the segments are termed elementary discourse units (EDUs), and a complete RST-style discourse analysis consists of building EDUs into a tree that spans the entire document. The tree edges are labeled with relations types, and nodes are categorized by their nuclearity (roughly, importance). RST segmentation is often regarded as a solved problem because automated segmenters achieve high performance (F1=94.3) on a task with high inter-annotator agreement (kappa=0.92) \cite{Wang:2018,Carlson:2001}.
In fact, many RST parsers do not include a segmenter and simply evaluate on gold EDUs. However, numerous studies have shown errors in segmentation are a primary bottleneck for accurate discourse parsing \cite{Soricut:2003,Fisher:2007,Joty:2015,Feng:2015}. Notably, even when using a top-performing segmenter, results degrade by 10\% on the downstream tasks of span, nuclearity and relation labeling when using predicted instead of gold EDUs \cite{Feng:2015}.

\begin{table}[t]
    \centering
    \begin{tabular}{p{7.3cm}}
    \toprule
    \textcolor{blue}{{\textbf{[}}}Patients were excluded{\textcolor{blue}{\textbf{]}}}
\textcolor{blue}{{\textbf{[}}}if they had any other major Axis I psychiatric disorder, any medical or neurological disorder{\textcolor{blue}{\textbf{]}}}
\textcolor{blue}{{\textbf{[}}}that could influence the diagnosis or treatment of depression,{\textcolor{blue}{\textbf{]}}}
\textcolor{blue}{{\textbf{[}}} any condition other than depression{\textcolor{blue}{\textbf{]}}}
\textcolor{blue}{{\textbf{[}}}that was not on stable treatment for at least the past one month,{\textcolor{blue}{\textbf{]}}}
\textcolor{blue}{{\textbf{[}}}any condition{\textcolor{blue}{\textbf{]}}}
\textcolor{blue}{{\textbf{[}}}that could pose a health risk during a clinical trial,{\textcolor{blue}{\textbf{]}}}
\textcolor{blue}{{\textbf{[}}}and any clinically significant abnormality or disorder{\textcolor{blue}{\textbf{]}}}
\textcolor{blue}{{\textbf{[}}}that was newly detected during the baseline assessments.{\textcolor{blue}{\textbf{]}}}  \\
    \bottomrule
    \end{tabular}
    \vspace{-0.6em}
    \caption{An example sentence from the novel \textsc{MEDICAL} corpus, with EDUs annotated in square brackets.}
    \label{tab:corpus_example}
\end{table}

Separately, all available discourse segmenters are trained on news, and their ability to generalize to other domains, such as medical text, has not been well-studied. In our work, we focus on the medical domain because it has garnered cross-disciplinary research interest with wide-reaching applications. For example, the Biomedical Discourse Relation Bank was created for PDTB-style discourse parsing of biomedical texts \cite{Prasad:2011}, and has been used to analyze author revisions and causal relations \cite{Zhang:2016,Marsi:2014}.


This work studies discourse segmentation in the medical domain. In particular, we: (1) seek to identify difficulties that news-trained segmenters have on medical; (2) investigate how features of the segmenter impact the type of errors seen in medical; and (3) examine the relationship between annotator agreement and segmenter performance for different types of medical data.

To this end, we present the first small-scale medical corpus in English, annotated by trained linguists (sample in Table \ref{tab:corpus_example}). We evaluate this corpus with three RST segmenters, finding an expected gap in the medical domain. We perform a detailed error analysis that shows medical-specific punctuation is the largest source of errors in the medical domain, followed by different word usage in syntactic constructions which are likely caused by news-derived word embeddings. Second, by comparing segmenters which use word embeddings versus syntax trees, we find access to parsed trees may not be helpful in reducing syntactically-resolvable errors, while an improved tokenizer would provide small benefits. Third, we note patterns between humans and segmenters where both perform better on extremely short texts and worse on those with more complex discourse. 

We conclude with suggestions to improve the segmenter on the medical domain and recommendations for future annotation experiments.

Our contributions in this work are two-fold: a high-quality small-scale corpus of medical documents annotated with RST-style discourse segments; a quantitative and qualitative analysis of the discourse segmentation errors in the medical domain that lays the groundwork for understanding both the strengths and limits of existing RST segmenters, and the next concrete steps towards a better segmenter for the medical domain.

\section{Related Work}
\noindent {\bf Corpora in non-news domains.}
The seminal RST resource, the RST Discourse Treebank (RST-DT) \cite{Carlson:2001}, consists of news articles in English. With the wide adoption of RST, corpora have expanded to other languages and domains. Several of these corpora include science-related texts, a domain that is closer to medical, but unfortunately also use segmentation guidelines that differ sometimes considerably from RST-DT\footnote{A future direction of research could revisit this domain of science if the differing segmentation schemes are adequately resolved in the forthcoming shared task of the Discourse Relation Parsing and Treebanking 2019 workshop.} (research articles in Basque, Chinese, English, Russian, Spanish \cite{Iruskieta:2013,Cao:2017,Zeldes:2017,Yang:2018,Toldova:2017,Dacunha:2012}; encyclopedias and science news web pages in Dutch \cite{Redeker:2012}). Specifically in the medical domain, only two corpora exist, neither of which are in English. \newcite{Dacunha:2012} annotate a small corpus of Spanish medical articles, and the RST Basque Treebank \cite{Iruskieta:2013} includes a small set of medical article abstracts. Our work aims to fill this gap by creating the first corpus of RST-segmented medical articles in English. Unlike several other works, we include all parts of the article, and not just the abstract.

\medskip
\noindent {\bf Segmenters in non-news domains.}
While corpora have expanded to other domains, most automated discourse segmenters remain focused (and trained) on news. An exception is the segmenter in \newcite{Braud:2017} which was trained on different domains for the purpose of developing a segmenter for under-resourced languages. However, they make the simplifying assumption that a single corpus represents a single (and distinct) domain, and do not include the medical domain. In this work, we study the viability of using news-trained segmenters on the medical domain.


\begin{table}[t]
\centering
\small
\begin{tabular}{ccccc}
\toprule
Corpus & \#docs  & \#tokens     & \#sents     & \#EDUs    \\ 
\midrule
\textsc{RST-DT small} &11  &4009  &159  &403  \\
                                  \textsc{Medical} & 11 &3356  &169  &399\\             \bottomrule
\end{tabular}
\vspace{-0.6em}
\caption{Corpus statistics.}
\label{tab:cross_domain_corpus}
\end{table}

\section{Corpus Creation}

\noindent{\bf Medical Corpus.}
 The \textsc{Medical} corpus consists of 2 clinical trial reports from PubMed Central, randomly selected for their shorter lengths for ease of annotation. We expect the language and discourse to be representative of this domain, despite the shorter length. As a result of the smaller size, we hypothesize annotator agreement and segmenter performance numbers may be somewhat inflated, but we nevertheless expect the nature of the errors to be the same. We divide the reports into their corresponding sections, treating each section as a separate document, resulting in 11 labeled documents. We chose to analyze sections individually instead of an entire report because moving to larger units typically yields arbitrary and uninformative analyses \cite{Taboada:2006}. XML formatting was stripped, and figures and tables were removed. The sections for \textit{Acknowledgements}, \textit{Competing Interests}, and \textit{Pre-publication History} were not included. 
 
 For comparison with the \textit{News} domain, we created \textsc{RST-DT-SMALL} by sampling an equal number of Wall Street Journal articles from the ``Test'' portion of the RST-DT that were similar in length to the medical documents. The corpus statistics are summarized in Table \ref{tab:cross_domain_corpus}.

\medskip
\noindent{\bf Annotation Process.} The annotation process was defined to establish a high-quality corpus that is consistent with the gold-segmented RST-DT. Two annotators participated: a Linguistics graduate student (the first author), and a Linguistics undergraduate (the second author). To train on the task and to ensure consistency with RST-DT, the annotators first segmented portions of RST-DT. During this training phase, they also discussed annotation strategies and disagreements, and then consulted the gold labels. In the first phase of annotation on the medical data, the second author segmented all documents over a period of three months using the guidelines compiled for RST-DT \cite{Carlson:2001b} and with minimal guidance from the first author. In the second phase of annotation, all documents were re-segmented by both annotators, and disagreements were resolved by discussion.

\medskip
\noindent{\bf Agreement.} Annotators achieved on average a high level of agreement for identifying EDU boundaries with kappa=0.90 (averaged over 11 texts). However, we note that document length and complexity of the discourse influence this number. On a document of 35 tokens, the annotators exhibited perfect agreement. For the \textit{Discussion} sections that make more use of discourse, the average agreement dropped to 0.84. The lowest agreement is 0.73 on a \textit{Methods} section, which had more complex sentences with more coordinated sentences and clauses, relative clauses and nominal postmodifiers (as discussed in Section \ref{sec:error_types}, these syntactic constructions are also a source of error for the automated segmenters).

\section{Experiment}
We automatically segment the documents in \textsc{RST-DT SMALL} and \textsc{MEDICAL} using three segmenters: (1) \textsc{DPLP}\footnote{https://github.com/jiyfeng/DPLP} uses features from syntactic and dependency parses for a linear support vector classifier; (2)\textsc{Two-pass} \cite{Feng:2014} is a CRF segmenter that derives features from syntax parses but also uses global features to perform a second pass of segmentation; (3) \textsc{Neural} \cite{Wang:2018} is a neural BiLSTM-CRF model that uses ELMo embeddings \cite{Peters:2018}. We choose these segmenters because they are widely-used and publicly available (most RST parsers do not include a segmenter). \textsc{DPLP} has been cited in several works showing discourse helps on different NLP tasks \cite{Bhatia:2015}. \textsc{Two-pass}, until recently, achieved SOTA on discourse segmentation when using parsed (not gold) syntax trees. \textsc{Neural} now holds SOTA in RST discourse segmentation. We evaluate the segmenter's ability to detect all EDU boundaries present in the gold data (not just intra-sentential) using the metrics of precision (P), recall (R) and F1.

The \textsc{DPLP} and \textsc{two-pass} segmenters, both of which employ the Stanford Core NLP pipeline \cite{Manning:2014}, were updated to use the same version of this software (2018-10-05).


\begin{table}[t]
\centering
\begin{tabular}{cllll}
\toprule
\multicolumn{1}{l}{\textsc{RST Seg}} & \textsc{Domain}  & \textsc{F1}     & \textsc{P}     & \textsc{R}    \\ 
\midrule
\multirow{2}{*}{\textsc{DPLP}}              & \textit{News}    & 82.56 & 81.75 & 83.37 \\
& \textit{Medical} & 75.29 & 78.69 & 72.18 \\
\midrule
\multirow{2}{*}{\textsc{two-pass}}              & \textit{News}    & 95.72 & \underline{97.19} & 94.29 \\
& \textit{Medical} & 84.69 &86.23 & 83.21 \\
\midrule
\multirow{2}{*}{\textsc{Neural}}              & \textit{News}    & \underline{97.32} & 95.68 & \underline{99.01} \\
& \textit{Medical} & \textbf{91.68} & \textbf{94.86} & \textbf{88.70} \\\bottomrule
\end{tabular}
\vspace{-0.6em}
\caption{F1, precision (P) and recall (R) of RST discourse segmenters on two domains (best numbers for \textit{News} are underlined, for \textit{Medical} are bolded).}
\label{tab:cross_domain_f1}
\end{table}

\begin{table*}[]
  \hfill
  \begin{minipage}{\textwidth}
  \centering
  \small
\begin{tabular}{lp{6.3cm}p{6.3cm}}
\toprule
\textsc{Error Type}    & \textsc{Predicted}     & \textsc{Gold} \\
\midrule
\multicolumn{1}{l} {amb. lexical cue} &{[}our performance\textcolor{red}{\bf {]}{[}}since the buy - out makes it imperative{]} &{[}our performance since the buy - out makes it imperative{]}\\
infinitival ``to'' &{[}the auto giants will move quickly\textcolor{red}{\bf{]}{[}}to buy up stakes{]} &{[}the auto giants will move quickly to buy up stakes{]}\\
correct &{[}you attempt to seize assets\textcolor{red}{\bf{]}{[}}related to the crime{]} &{[}you attempt to seize assets related to the crime{]}\\
\midrule
\multicolumn{1}{l}{tokenization} & {[}as identified in clinical \textcolor{red}{\bf trials.\{8-11\}It{]}{[}}is noteworthy{]} & {[}as identified in clinical trials .{]}{[}\{ 8-11 \}{]}{[}It is noteworthy{]}       \\ 
end emb. EDU & {[}Studies{]}{[} confined to medical \textcolor{red}{\bf professionals have} shown{]}                                           & {[}Studies{]}{[} confined to medical professionals{]}{[}have shown{]}                                                    \\ punctuation & {[}the safety of placeboxetine{]}{[}( PB \textcolor{red}{\bf) hydrochloride} capsules{]}& {[}the safety of placeboxetine{]}{[}( PB ){]}{[} hydrochloride capsules{]}\\
\bottomrule
\end{tabular}
\vspace{-0.6em}
  \end{minipage}
\caption{Examples of the most frequent segmentation error types with the erroneous EDU boundaries highlighted in red for \textit{News} (top) and \textit{Medical} (bottom) with predicted and gold EDU boundaries in square brackets (square brackets for citations are changed to curly brackets to avoid confusion). For \textit{News}, the boundaries are inserted incorrectly (false positives) and for \textit{Medical} they are omitted incorrectly (false negatives).}
\label{tab:cross_domain_errors}
\end{table*}

\section{Results}
Table \ref{tab:cross_domain_f1} lists our results on \textit{News} and \textit{Medical} for correctly identifying EDU boundaries using the three discourse segmenters. As expected, the \textit{News} domain outperforms the \textit{Medical} domain, regardless of which segmenter is used. In the case of the \textsc{DPLP} segmenter, the gap between the two domains is about 7.4 F1 points. Note that the performance of \textsc{DPLP} on \textit{News} lags considerably behind the state of the art (-14.76 F1 points). When switching to the \textsc{two-pass} segmenter, the performance on \textit{News} increases dramatically (+13 F1 points). However, the performance on \textit{Medical} increases by only 3.75 F1 points. Thus, large gains in \textit{News} translate into only a small gain in \textit{Medical}. The \textsc{neural} segmenter achieves the best performance on \textit{News} and is also able to more successfully close the gap on \textit{Medical}, with only a 5.64 F1 difference, largely attributable to lower recall.

\section{Error Analysis}
We perform an error analysis to understand the segmentation differences between domains and between segmenters.

\subsection{Error Types}
\label{sec:error_types}
We first group errors of the best-performing \textsc{neural} segmenter into \emph{error types}. Here we discuss the most frequent types in each domain and give examples of each in Table \ref{tab:cross_domain_errors} with the predicted and gold EDU boundaries.

\medskip
\noindent {\bf ambiguous lexical cue} Certain words (often discourse connectives) are strongly indicative of the beginning of an EDU, but are nonetheless ambiguous because of nuanced segmentation rules. In the Table \ref{tab:cross_domain_errors} example, the discourse connective ``since'' typically signals the start of an EDU (e.g., in the RST discourse relations of \textit{temporal} and \textit{circumstance}), but is not a boundary in this case because there is no verbal element. Other problematic words include ``that'', signalling relative clauses (often, but not always treated as embedded EDUs), and ``and'' which may indicate a coordinated sentence or clause (treated as a separate EDU) but also a coordinated verb phrase (not a separate EDU). Note this phenomenon is different from distinguishing between discourse vs. non-discourse usage of a word, or sense disambiguation of a discourse connective as studied in \citet{Pitler:2009}.

\noindent {\bf infinitival ``to''} The syntactic construction of \texttt{to+verb} can act either as a verbal complement (treated as the same EDU) or a clausal complement (separate EDU). In the Table \ref{tab:cross_domain_errors} example, the infinitival ``to buy'' is a complement of the verb ``move'' and should remain in the same EDU, but the segmenter incorrectly segmented it.

\noindent {\bf tokenization} This error type covers cases where the tokenizer fails to detect token boundaries, specifically punctuation. These tokenization errors lead to downstream segmentation errors since punctuation marks, often markers of EDU boundaries, are entirely missed when mangled together with their neighboring tokens, as in `trials.[8-11]It' in Table \ref{tab:cross_domain_errors}. 

\noindent {\bf punctuation} This error occurs when parentheses and square brackets are successfully tokenized, but the segmenter fails to recognize them as EDU boundaries. This error is expected for square brackets, as they do not occur in RST-DT, but frequently appear in the \textit{Medical} corpus for citations. It is not clear why the segmenter has difficulty with parentheses as in the Table \ref{tab:cross_domain_errors} example ``(PB)'', since they do occur in \textit{News} and further almost invariably mark an EDU boundary. 

\noindent {\bf end of embedded EDU} An embedded EDU breaks up a larger EDU and is typically a relative clause or nominal postmodifier with a verbal element.\footnote{For a more complete definition, see the tagging manual.} While the segmenter is good at identifying the beginning of an embedded EDU, it often fails to detect the end. An embedded EDU such as the one listed in Table \ref{tab:cross_domain_errors} can be clearly identified from a syntactic parse: the verbal element `have shown' attaches to the subject `Studies' and not the nominal postmodifier as predicted by the segmenter.

\noindent {\bf correct} This category describes cases where we hypothesize the annotator made a mistake and the segmenter is correct. In the Table \ref{tab:cross_domain_errors} example, the nominal postmodifier with non-finite clause ``related to the crime'' is an embedded EDU missed by annotators.

\begin{figure*}
\centering
\begin{subfigure}{.5\textwidth}
  \centering
  \includegraphics[width=1\linewidth]{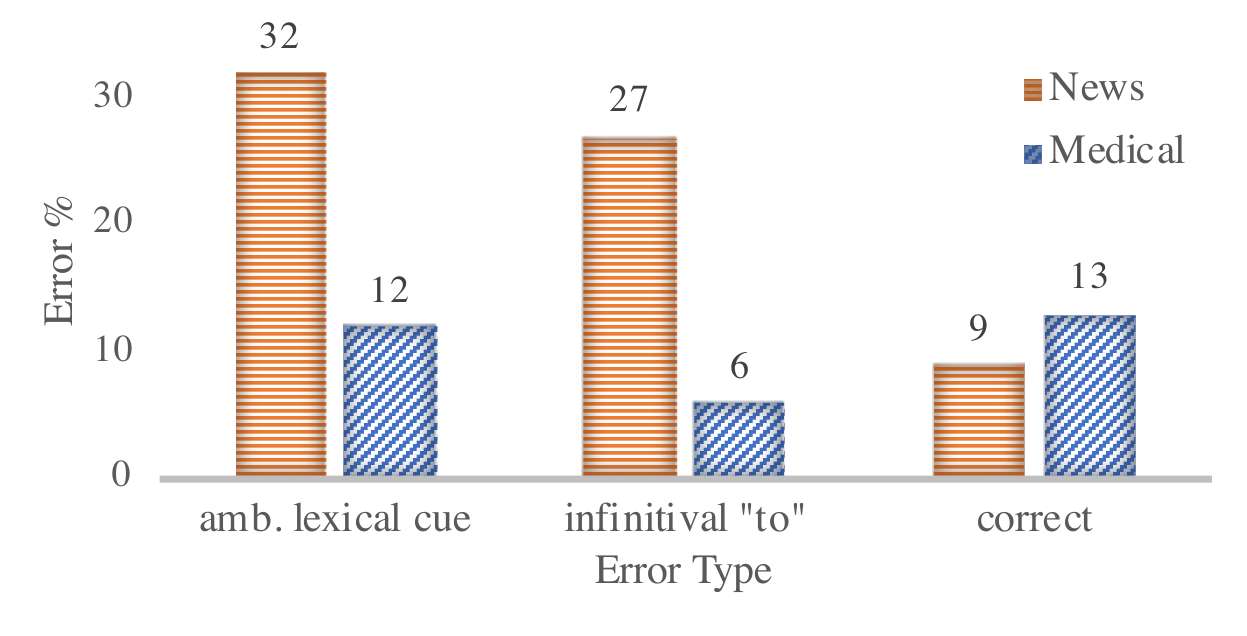}
  \vspace{-2.2em}
  \caption{most frequent in \textit{News}}
  \label{fig:errors_news}
\end{subfigure}%
\begin{subfigure}{.5\textwidth}
  \centering
  \includegraphics[width=1\linewidth]{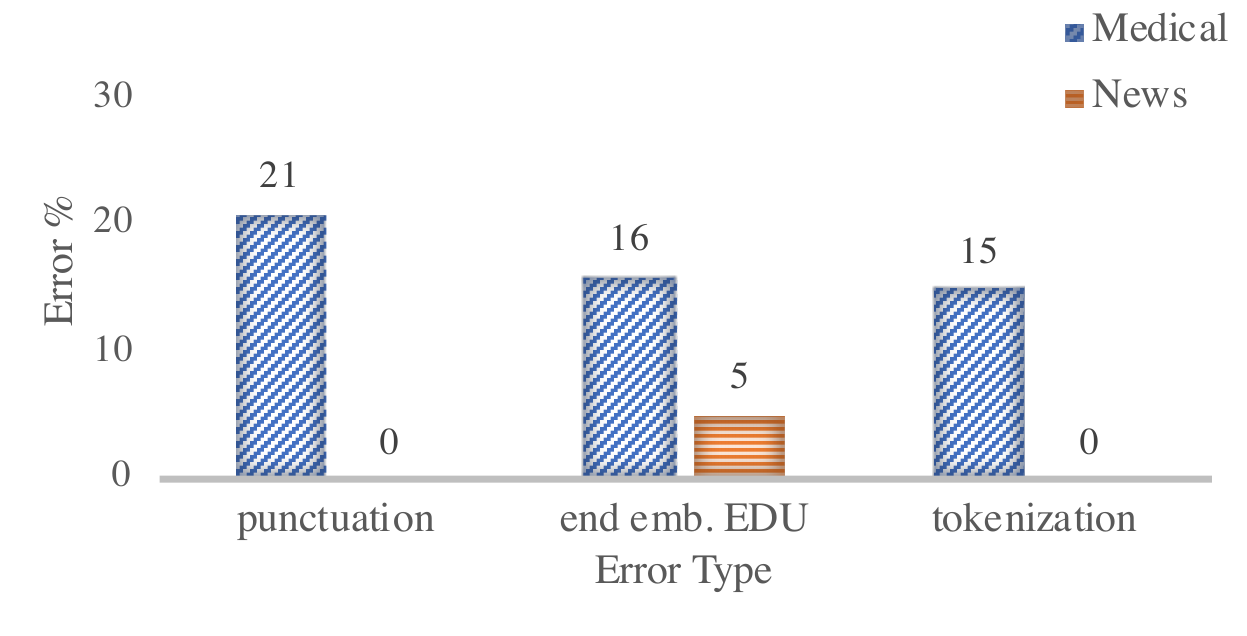}
  \vspace{-2.2em}
  \caption{most frequent in \textit{Medical}}
  \label{fig:errors_medical}
\end{subfigure}
\caption{Most frequent segmentation error types by domain, using the best discourse segmenter.}
\label{fig:errors}
\end{figure*}

\subsection{Errors between domains}

In Figure \ref{fig:errors}, we compare the distribution of the most frequent error types in \textit{News} (left) and the most frequent in \textit{Medical} (right). 

In \textit{News} Figure \ref{fig:errors_news}, the errors are mostly false positives where the segmenter incorrectly inserts boundaries before \textit{ambiguous lexical cues}, and before \textit{infinitival ``to''} clauses (that are verbal complements).  Interestingly, \newcite{Braud:2017} found the tricky distinction of clausal vs. verbal infinitival complements to also be a large source of segmentation errors. These two error types also occur in \textit{Medical}, though not as frequently, in part because the \texttt{to+verb} construction itself occurs less in the medical corpus. The third category of \textit{correct} consists mostly of cases where the segmenter correctly identified an embedded EDU missed by annotators, illustrating both the difficulty of annotation even for experts and the usefulness of an automated segmenter for both in-domain and out-of-domain data since this error type is attested in both domains.

In \textit{Medical} Figure \ref{fig:errors_medical}, we first note a stark contrast in distribution between the domains.  The error types most frequent in \textit{Medical} are hardly present in \textit{News}; that is, errors in the \textit{Medical} domain are often exclusive to this domain. The errors are mostly false negatives where the segmenter fails to detect boundaries around medical-specific use of \textit{punctuation} marks, including square brackets for citations and parentheticals containing mathematical notations, which are entirely absent in \textit{News}. The segmenter often misses the \textit{end of embedded EDU}s, and more frequently than in \textit{News}. The difference in this syntactically-identifiable error points to a gap in the embedding space for words signalling relative clauses and nominal postmodifiers. Given that ELMo embeddings have been shown to capture some syntax \cite{Tenney:2019}, we recommend using PubMed-trained ELMo embeddings.\footnote{This option is viable once the \textsc{Medical} corpus is expanded to a large enough size for training.} One may further hypothesize that adding syntactic parses to the segmenter would help, which we explore in Section \ref{sec:errors_segmenters}. The third error of \textit{tokenization} occurs mainly around square brackets (citations), and this specific token never occurs in \textit{News}. 

\begin{figure}
    \centering
    \includegraphics[width=\columnwidth]{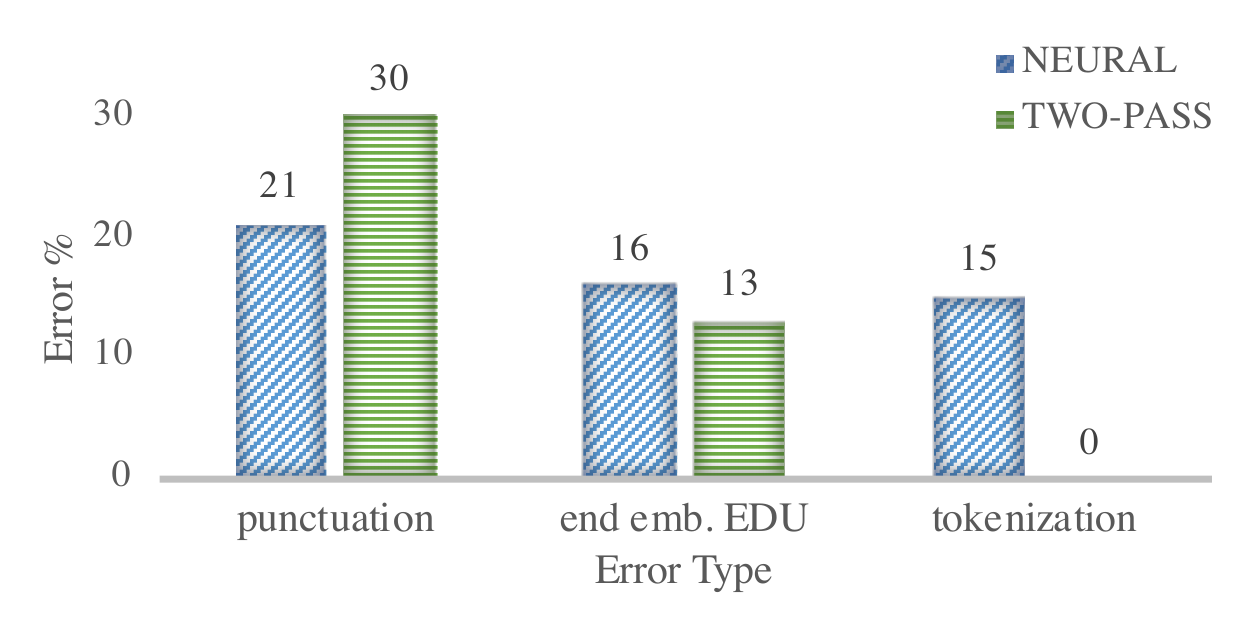}
    \vspace{-2.2em}
    \caption{Distribution of the most frequent error types on \textit{Medical} when using the \textsc{Neural} and \textsc{Two-pass} segmenter.}
    \label{fig:errors_syntax}
\end{figure}

\subsection{Errors between segmenters}
\label{sec:errors_segmenters}
The rules of discourse segmentation rely heavily on syntax. Most discourse segmenters include syntax parse trees with the notable exception of the \textsc{Neural} segmenter. While this is the best-performing segmenter, we question whether it could be improved further if it had access to syntax trees. We probe this question by comparing the \textsc{Neural} segmentation errors with those found in \textsc{Two-pass}, which does use syntax trees. 

Figure \ref{fig:errors_syntax} illustrates the proportion of error types using the two segmenters. Although \textsc{two-pass} makes use of syntax trees, the frequency of the syntactically-identifiable \textit{end of embedded EDU} error type is only slightly lower. Because we do not have gold trees, it is also possible the news-trained parser performs poorly on medical and leads to downstream errors. We visually inspect the parse trees for these error cases and find the syntactic clause signaling the embedded EDU is correctly parsed in half the cases. Thus, bad parse trees contribute only partially to this error, and we suspect better trees may not provide much benefit. This finding is consistent with the little help dependency trees provided for cross-lingual discourse segmentation in \newcite{Braud:2017b}.

We further note the tokenizer for \textsc{two-pass} makes no errors on the medical data, but conversely has a higher proportion of \textit{punctuation} errors. This pattern suggests improving the tokenizer of the \textsc{neural} segmenter may simply shift errors from one type to another. To test this hypothesis, we use pre-tokenized text and find roughly half the errors do shift from one type to the other, but the other half is correctly labeled. That is, performance actually improves, but only slightly (F1=+0.36, P=+0.50, R=+0.24).

\subsection{Errors between annotators and segmenters}
Here we compare the level of annotator agreement with the performance of the \textsc{neural} segmenter. In Table \ref{tab:anno_f1}, we see that both humans and the model do well on extremely short texts (\textit{Summary}). However, high agreement does not always translate to good performance. The \textit{Introduction} section is straightforward for the annotators to segment, but this is also where most citations occur, causing the segmenter to perform more poorly. Earlier, we had noted the \textit{Discussion} section was the hardest for annotators to label because of the more complex discourse. These more ambiguous syntactic constructions also pose a challenge for the segmenter, with lower performance than most other sections. 

\begin{table}[]
    \centering
    \begin{tabular}{llll}
    \toprule
         \textsc{Section}&  \textsc{Kappa} & \textsc{F1} & \textsc{\#tokens}\\
         \midrule
         Summary & 1.00 &100 & 35 \\
         Introduction & 0.96 &86.58 & 258 \\
         Results & 0.93 & 91.74 & 354 \\
         Abstract & 0.89 & 95.08 & 266 \\
         Methods & 0.86 & 92.99 & 417 \\
         Discussion & 0.84 & 89.03 & 365 \\
         \bottomrule
    \end{tabular}
    \vspace{-0.6em}
    \caption{Average inter-annotator agreement per section, ordered from highest to lowest, the corresponding average F1 of the \textsc{neural} segmenter, and number of tokens (there are 2 documents per section, except 1 for Summary).}
    \label{tab:anno_f1}
\end{table}

\section{Next Steps}
\label{sec:next_steps}
Based on our findings, we propose a set of next steps for RST discourse analysis in the medical domain. A much faster annotation process can be adopted by using the \textsc{neural} segmenter as a first pass. Annotators should skip extremely short documents and instead focus on the more challenging \textit{Discussion} section. During training, we recommend using medical-specific word embeddings and preprocessing pipeline.\footnote{\url{https://allennlp.org/elmo},\url{https://allenai.github.io/scispacy}} 
Addressing even one of these issues may yield a multiplied effect on segmentation improvements as the \textit{Medical} domain is by nature highly repetitive and formulaic.

However, a future avenue of research would be to first understand what impact these segmentation errors have on downstream tasks. For example, using RST trees generated by the lowest-performing \textsc{DPLP} parser nevertheless provides small gains to text categorization tasks such as sentiment analysis \cite{Ji:2017}. On the other hand, understanding the verb form, which proved to be difficult in the \textit{Medical} domain, has been shown to be useful in distinguishing text on experimental results from text describing more abstract concepts (such as background and introductory information), which may be a more relevant task than sentiment analysis \cite{deWaard:2012}. 

\section{Conclusion}
As a first step in understanding discourse differences between domains, we analyze the performance of three discourse segmenters on \textit{News} and \textit{Medical}. For this purpose, we create a first, small-scale corpus of segmented medical documents in English. All segmenters suffer a drop in performance on \textit{Medical}, but this drop is smaller on the best \textit{News} segmenter. An error analysis reveals difficulty in both domains for cases requiring a fine-grained syntactic analysis, as dictated by the RST-DT annotation guidelines. This finding suggests a need for either a clearer distinction in the guidelines, or more training examples for a model to learn to distinguish them. In the \textit{Medical} domain, we find that differences in syntactic construction and formatting, including use of punctuation, account for most of the segmentation errors. We hypothesize these errors can be partly traced back to tokenizers and word embeddings also trained on \textit{News}. We finally compare annotator agreement with segmenter performance and find both suffer in sections with more complex discourse. Based on our findings, we have proposed (Section \ref{sec:next_steps}) a set of next steps to expand the corpus and improve the segmenter.

\section*{Acknowledgments}
We thank the anonymous reviewers for their helpful feedback. The first author was supported by the National Science Foundation Graduate Research Fellowship Program under Grant No. 2017247409. Any opinions, findings, and conclusions or recommendations expressed in this material are those of the authors and do not necessarily reflect the views of the National Science Foundation.

\bibliography{naaclhlt2019}
\bibliographystyle{acl_natbib}

\end{document}